\documentclass[letterpaper]{article}
\usepackage[final]{nips_2016}
\usepackage{times}
\usepackage{helvet}
\usepackage{courier}
\usepackage{latexsym}
\usepackage{url}
\usepackage[draft]{hyperref}
\usepackage{amssymb}
\usepackage{amsmath}
\DeclareMathOperator{\E}{\mathbb{E}}
\usepackage{floatrow}
\usepackage{graphicx}
\graphicspath{ {images/} }
\usepackage[font=small]{caption}
\usepackage{wrapfig}
\usepackage{float}

\def\figref#1{Fig.~\ref{#1}}

\def\tabref#1{Table~\ref{#1}}

\usepackage[colorinlistoftodos]{todonotes} %For adding todo

\title{Message Passing Multi-Agent GANs}

\author{
Arnab Ghosh\thanks{Equal Contribution} , Viveka Kulharia$^\ast$, Vinay Namboodiri\\
$$IIT Kanpur\\ 
{\tt \{arnabghosh93,vivekakulharia\}@gmail.com}, \\ {\tt vinaypn@iitk.ac.in} 
}

\begin{document}

\maketitle

\begin{abstract}
Communicating and sharing intelligence among agents is an important facet of achieving Artificial General Intelligence. 
As a first step towards this challenge, we introduce a novel framework for image generation: Message Passing Multi-Agent Generative Adversarial Networks (MPM GANs). While GANs have recently been shown to be very effective for image generation and other tasks, these networks have been limited to mostly single generator-discriminator networks. We show that we can obtain multi-agent GANs that communicate through message passing to achieve better image generation. The objectives of the individual agents in this framework are two fold: a co-operation objective and a competing objective. The co-operation objective ensures that the message sharing mechanism guides the other generator to generate better than itself while the competing objective encourages each generator to generate better than its counterpart. We analyze and visualize the messages that these GANs share among themselves in various scenarios. We quantitatively show that the message sharing formulation serves as a regularizer for the adversarial training. 
Qualitatively, we show that the different generators capture different traits of the underlying data distribution.
\end{abstract}

\section{Introduction}

Unsupervised learning has emerged as one of the most important facets of Machine Learning research. With the advent of Generative Adversarial Networks (GANs) (\cite{goodfellow2014generative}) it has become possible to harness large amounts of unlabeled data in the form of a generative model which can make extremely plausible images (\cite{radford2015unsupervised}). If we are to target superhuman intelligence, we have to create networks that not only learn from large quantities of data but also interact among themselves in order to learn from each other or even compete with each other. Through this work we obtain one of the first approaches towards engaging multiple agents towards learning of deep unsupervised representations. Note that very recently multiple agents have been explored by \cite{sukhbaatar2016learning} and \cite{foerster2016learning} where they employ a Deep Reinforcement based formulation in order to achieve a shared utility.

Generative Adversarial Networks have recently seen applications in Image Inpainting (\cite{pathak2016context}), Interactive Image Generation from just a few brushstrokes (\cite{zhu2016generative}), Image Super Resolution (\cite{ledig2016photo}) and Abstract Reasoning Diagram Generation (\cite{ghosh2017contextual}). GANs have been augmented in several ways to extract structure out of the representations most notably by \cite{chen2016infogan},
\cite{liu2016coupled} and \cite{dumoulin2016adversarially}. While our Multi-Agent Generator based framework has the elegance to be applicable in most of the above applications, we demonstrate its application in the task of unsupervised image generation. 

\begin{figure}[H]
    \centering
    \includegraphics[scale=0.5]{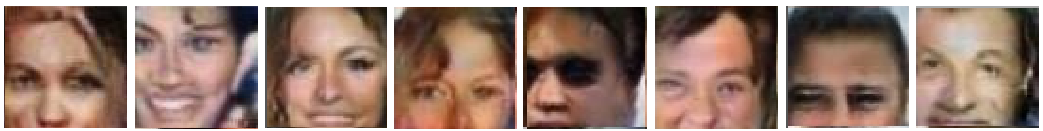}
	\caption{Generations of Generator 1 with uniform(-1,1) noise distribution with conditioned message passing. It captures detailed facial expression.}
   \label{fig:diffnoiseG1}
\end{figure}

\begin{figure}[H]
    \centering
    \includegraphics[scale=0.5]{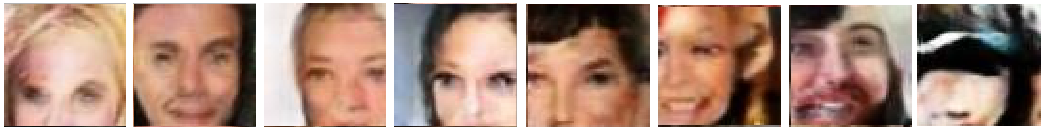}
	\caption{Generations of Generator 2 with N(0,1) noise distribution with conditioned message passing. It captures smooth features of facial expression.}
   \label{fig:diffnoiseG2}
\end{figure}

\begin{figure}[h]
    \centering
    \includegraphics[scale=0.98]{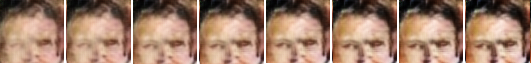}
	\caption{Generations look as if showing the process of artist creation.}
    \label{fig:artistic}
\end{figure}

Our work bears close resemblance to the work on Adversarial Neural Cryptography (\cite{abadi2016learning}) where the Cryptographic System is automatically learned based on the varying objectives of the three agents Alice, Bob and Eve. Our conceding and competing objectives are based on these ideas. Multi-Agent Systems with message passing were first employed by \cite{foerster2016learning} and our model of harnessing the messages that are received from the other generator is based on similar ideas. \cite{lazaridou2016multi} introduced a message passing model between the different agents that are forcibly made to co-operate via the introduction of a bottleneck and the clustering of the image features show that messages for images of same category are usually same.

This work presents one of the first forays into this subject and in a more traditional Deep Unsupervised Learning setting rather than in a Deep Reinforcement Setting where the reward structure is discrete and training becomes slightly more difficult. In this work we present a setting of Multi-Generator based Generative Adversarial Networks with a competing objective function which promotes the two generators to compete among themselves apart from trying to maximally fool the discriminator. We also analyze a conceding objective which tries to promote the other generator to be better than itself. We also introduce a message passing model in order to make the generators aware of the generations the other generator is targeting and hence learn to generate better images. 

With the message passing model, a fact emerged that a bottleneck has to be added in order to make the message generator actually learn meaningful representations of the messages. Hence we demonstrate the performance of the message passing model in presence of three bottlenecks. The first being the two generators being passed samples from different noise distributions, namely one of them was provided with samples from Normal(0,1) and the other one was passed samples from Uniform(-1,1). The message passing model was also analyzed with the two objectives introduced as competing objective and the conceding objective to understand the message and the generations that each of the networks produce in such situations.

The models yielded some interesting results as seen from \figref{fig:diffnoiseG1} and \figref{fig:diffnoiseG2} without any explicit formulation one of the generators was generating images with much more facial detail while the other generator was generating images with the overall content with even obscure objects as the last image in \figref{fig:diffnoiseG2} where a  woman wearing a cap with her eyes covered by the cap is depicted. An interesting observation was in \figref{fig:artistic} where the message interpolation results between the 2 generators showed the process an artist takes for an artistic creation.

In summary our main contributions in the paper are:
\begin{itemize}
\item Presenting a novel framework of Multi-Agent GANs that comprises of multiple generators.
\item Introducing an objective which promotes competition among the generators and another objective which tries to make the other generator better than the current generator.
\item Introducing a novel message passing model, with the messages being passed between the generators in order to better explore the modes in the distribution.
\end{itemize}

\section{Related Work}
Unsupervised Learning with Generative Models have made immense progress within a remarkably short time, most notably being pioneered by 2 major directions Variational Autoencoders (\cite{kingma2013auto}) and Generative Adversarial Networks (\cite{goodfellow2014generative}). Efforts have been made for the unification of the 2 methods using Adversarial Autoencoders (\cite{makhzani2015adversarial}). Since the Variational Autoencoder based models are based on a Maximum Likelihood based objective hence some of the modes may remain unexplored.

\paragraph{GAN} Generative Adversarial Networks (GANs) have received tremendous interest in recent times especially after \cite{radford2015unsupervised} were able to show several interesting interpolation based generations and even arithmetic properties that exists in the latent space. Several applications such as video generation (\cite{vondrick2016generating}), Image manipulation (\cite{zhu2016generative}) and 3-D object generation (\cite{wu2016learning}) use GANs as the underlying generative model. Several variants of the GAN training objective have also been proposed in order to stabilize the training such as \cite{salimans2016improved} and \cite{arjovsky2016towards}.  Several objective functions have been proposed which minimize a divergence different from the Jensen  Shannon divergence as proposed by  \cite{goodfellow2014generative} for instance  \cite{nowozin2016f} experiment with various different divergences and show improved results.

\paragraph{Conditional GANs} Our technical approach is closely related to the Conditional GANs of \cite{mirza2014conditional} which generate images based on class specific information, \cite{reed2016generative} which condition the generation on the text, \cite{ghosh2017contextual} which condition the generation on all previous inputs via a RNN, \cite{chen2016infogan} which learn special representations of its latent variables for an interpretable conditional GAN based model. \cite{durugkar2016generative} also looked upon multi-agent GANs but their model was based on multiple discriminators rather than multiple generators and based on ensemble based principles rather than message passing based objective. \cite{liu2016coupled} learn a joint distribution of images by coupling a pair of GANs i.e. jointly training a pair of generator-discriminator such that some of the initial layers of generators have shared weights and similarly some of the last layers of the discriminators have shared weights. 
\paragraph{Message Passing Models and Co-operating Agents}
Belief propagation (\cite{weiss2001optimality}) based message passing had been one of the major learning algorithms employed as the principal training procedure in Probabilistic Graphical Models. The paradigm of co-operating agents has been looked upon in Game Theory (\cite{cai2011minmax}). \cite{foerster2016learning} and \cite{sukhbaatar2016learning} introduce formulations of co-operating agents with a message passing model and a common communication channel respectively. \cite{lazaridou2016multi} recently introduced a framework for networks to work co-operatively and introduce a bottleneck that forces the networks to pass messages which are even interpretable by humans.

\paragraph{Competing Agents}
Although the Generative Adversarial Networks \cite{goodfellow2014generative} is itself modeled as an adversarial game between 2 agents but with the advent of the competing objective even between the competing generators the generators start venturing into slightly different modes in the underlying noise space exploring greater modes of data. \cite{lee2016stochastic} is a work that incorporates competition between deep ensembles by passing the gradient to the best network. \cite{abadi2016learning} formulated a neural cryptography based framework in which Eve is an adversary and Alice and Bob work co-operatively in order to hide sensitive information from Eve.

\section{Models}
With the introduction of multiple generators, we add another set of objectives which helps us understand the dynamics of the system. We also introduce a version of Message Passing Generative Adversarial Networks with several variations which pass messages in order to make the generations of both better. The message passing model is augmented with several bottlenecks which encourage the generators to pass meaningful messages.  
\paragraph{Competing Objective}
The competing objective that we introduced is based on the principle that the Generators also compete with each other to get better scores for its generations from the Discriminator. The minimization objective function for the generator $G_1$ is:
\begin{equation}
\E_{z\sim p_z(z)}[log(1-D(G_1(z))) - f( D(G_1(z))-D(G_2(z)))]
\end{equation}

while the minimization objective function for generator $G_2$ is:
\begin{equation}
\E_{z\sim p_z(z)}[log(1-D(G_2(z))) - f( D(G_2(z))-D(G_1(z)))]
\end{equation}

where  f(x) = $max(x,0)$ so that the optimization objective for $G_1$ pushes it to get better scores from the Discriminator and vice versa for $G_2$.

\paragraph{Conceding Objective}
The principle behind the introduction of this objective is that the 2 generators try to guide each other in order to get better scores for its generations from the Discriminator. This model is similar in structure to the Competing Objective but the crucial difference is in the function used. Here, the minimization
objective function for the generator $G_1$ is:
\begin{equation}
\E_{z\sim p_z(z)}[log(1-D(G_1(z))) + f( D(G_1(z))-D(G_2(z)))]
\end{equation}

while the minimization
objective function for the generator $G_2$ is:
\begin{equation}
\E_{z\sim p_z(z)}[log(1-D(G_2(z))) + f( D(G_2(z))-D(G_1(z)))]
\end{equation}

where f(x) = $max(x,0)$ so that the generations of $G_2$ are better than that of $G_1$. 

\paragraph{Message Passing}
Our Message Passing model is based upon the principle that the messages passed between the 2 Generators will make the generators explore different subspaces of the image manifold and also provide better training for the discriminator as a regularization by introducing different types of images to the Discriminator.

\paragraph{Message Passing Model}
\begin{figure}[h]
    \centering
    \includegraphics[scale=0.25]{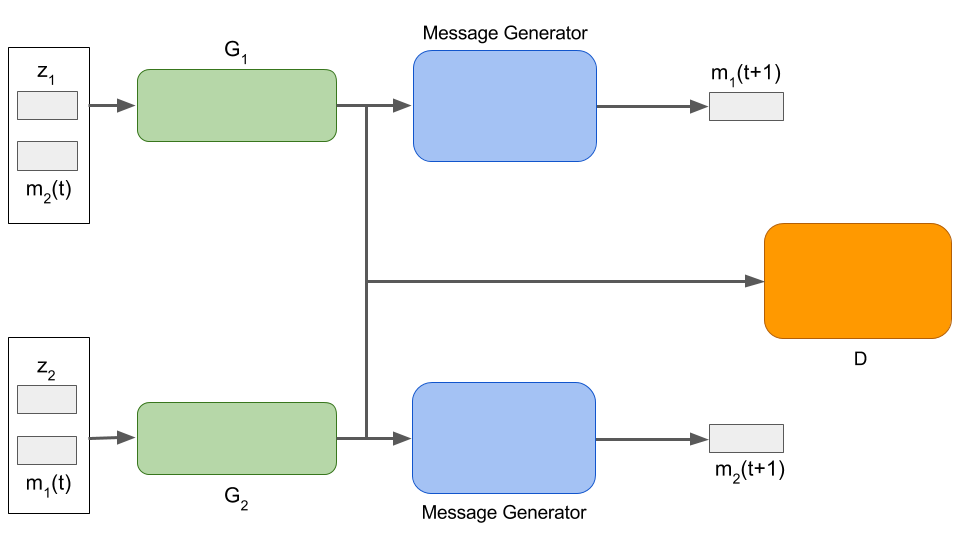}
	\caption{Model based on message passing without condition. The pair of Message Generators share the parameters between themselves.}
   \label{fig:msg_pass}
\end{figure}

Each Generator generates images conditioned upon the message that it receives from the other generator and the noise sampled from the noise distribution. After both the generators have generated their respective images, a common message generator with shared parameters takes the image as input and generates message and the message generated from each generator's image is passed to the other generator as a message in the next iteration. We also experimented with individual message generators for each generator but a common message generator works better because the messages are transferred between the 2 generators and meaningful messages can only be produced if the same network can gauge the generations by the 2 generators.

The minimization objective function for the generator $G_1$ is
\begin{equation}
\E_{x\sim [p_z(z), msg(G_2(z_2,m_1))]}[log(1-D(G_1(x)))]
\end{equation}
where x is composed of noise obtained from distribution $p_z$ and message passed by $G_2$. The message is initialized with same distribution as noise. $m_2$ is the message for the Generator G1 created by the message generator in the previous iteration.

Similarly, the minimization objective function for the generator $G_2$ is
\begin{equation}
\E_{x\sim [p_z(z), msg(G_1(z_1,m_2))]}[log(1-D(G_2(x)))]
\end{equation}

The discriminator is trained such that both the generations by $G_1$ and $G_2$ are labeled to be as fake by the discriminator.

\paragraph{Conditioned Message Passing Model}
\begin{figure}[h]
    \centering
    \includegraphics[scale=0.25]{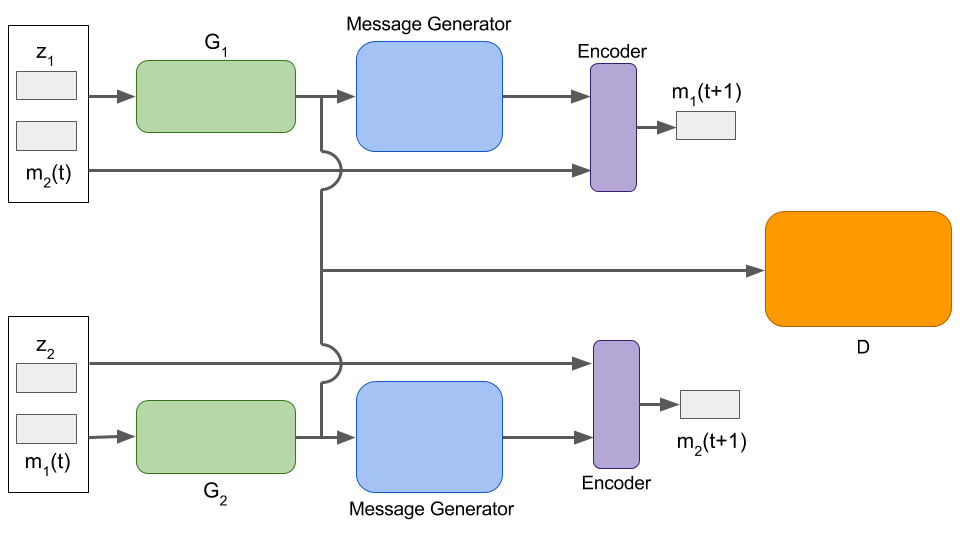}
	\caption{Model based on message passing with condition. Both the pairs of Message Generators and Encoders share the parameters between themselves.}
   \label{fig:cond_msg_pass}
\end{figure}
Each generated image is passed to Message Generator which creates an output. This output along with the generator's input is encoded using a multi-layer perceptron called Encoder to create the message. As the message is conditioned both on the generation and the input of the generator, the encoder can create much better messages as it knows what factors led to the generation. 
 
The objective being minimized by the generator $G_1$ is
\begin{equation}
\E_{x\sim [p_z(z), enc[msg(G_2(z_2,m_1)),z_2,m_1]]}[log(1-D(G_1(x)))]
\end{equation}
where x is composed of noise obtained from distribution $p_z$ and message passed by $G_2$. The message is initialized with same distribution as noise.
Analogously, the objective being minimized by the Generator $G_2$ is
\begin{equation}
\E_{x\sim [p_z(z), enc[msg(G_1(z_1,m_2)),z_1,m_2]]}[log(1-D(G_2(x)))]
\end{equation}

The message passing model is oblivious to the input that the generator received in order to generate the images and hence doesn't give good generations, on the other hand the conditioned message generation gives much better generations because the messages are also conditioned on the input and the output of the generators.

We consider three different bottlenecks in order to force the messages to be meaningful:

\paragraph{Different Noise Distributions}
The noise $z_1$ and $z_2$ that each of the Generators $G_1$ and $G_2$ get are sampled from different distributions. The principle behind the introduction of this bottleneck is that the generators would be able to master the modes in the 2 kinds of noise distributions and additionally that the messages will be forced to be different from mirroring the trivial noise distribution that was initially started off with. More concretely $z_1 \sim Uniform(-1,1)$ and $z_2 \sim Normal(0,1)$ was used for the training of the pair of generators.

\paragraph{Conceding Objective}
In order for the generators to co-operate and pass meaningful messages and make each other better we provide a model where the generators' objective function tries to make the other generator's generations get better scores from the discriminator and passes messages accordingly.

The objective function minimized by the generator $G_1$ is:
\begin{equation}
\E_{x\sim [p_z(z), enc[msg(G_2(z_2,m_1)),z_2,m_1]]}[log(1-D(G_1(x)))] + f(D(G_1(x))-D(G_2(x)))]
\end{equation}
and similar version for the generator $G_2$

\paragraph{Competing Objective}
In order to see the effects of the competing objective with the message passing model and whether rogue messages are passed in order to get better scores for its generations from the other discriminator, we provide a model of message passing GANs which compete with each other along with passing messages.

The structure is same as the message passing with condition. The objective function minimized by the generator $G_1$ is:
\begin{equation}
\E_{x\sim [p_z(z), enc[msg(G_2(z_2,m_1)),z_2,m_1]]}[log(1-D(G_1(x)))] - f(D(G_1(x))-D(G_2(x)))]
\end{equation}
and similar version for the generator $G_2$
\paragraph{Discriminator Objective}
In the simplest version (when generators don't pass messages) the objective function being maximized by the discriminator is:
\begin{equation}
\begin{split}
\E_{x\sim p_{data}(x)}[log(D(x)))] + \E_{z1\sim p_{z1}(z_1)}[log(1-D(G_1(z_1))))] \\+ \E_{z2\sim p_{z_2}(z2)}[log(1-D(G_2(z_2))))]
\end{split}
\end{equation}
When generators pass messages, only the input of the generators will change to include messages as well.

\section{Experimental Setup}

\paragraph{Model Architecture Details}
The Generator and the Discriminator's architecture was unaltered from \cite{radford2015unsupervised} while the only change was with the introduction of the message generator which has an almost identical architecture as the Discriminator but with the modification of changing the number of filters to the message dimension of the final output. On extensive experimentation with the different dimensions used for the message the best results were produced when the dimension of the message was 50.
The experiments done are:
\paragraph{Classification}
The representation of the image obtained by passing the real images through the discriminator as employed by \cite{radford2015unsupervised} was used alongside a novel feature representation enabled by our formulation of the message generator. The interesting aspect of the message generator is that it never got to see the real images, it just got to see the generated images by the 2 Generators and still when its feature representation is used it still gives interesting results. The dataset used for the classification examples is the Street View House Numbers Dataset \cite{goodfellow2013multi} which was used by \cite{radford2015unsupervised} and also \cite{salimans2016improved} for the evaluation of their techniques. Ablation studies were performed to identify the benefits from the discriminator representation and the message representation individually as well.

\paragraph{Clustering}
The celebrity dataset \cite{liu2015faceattributes} was used to partition the faces based on the type of hair into five categories: bald, black, brown, blond and gray. The images belonging to these partitions were passed through the Message Generator to get the representations of each of the images from the Message Generator. The representations are then reduced to 2 dimensions using T-SNE and then they are represented using the different colors. Somewhat meaningful clusters start emerging from this exercise.
 
\paragraph{Visualization}

With the introduction of the message passing mechanism, visualization was done by varying the messages or the noise in order to interpret the manifold learnt by the pair of Generators.

\begin{itemize}
\item \textbf{Message Interpolation:} Keeping the noise constant between the 2 generators we can understand the structure of the messages learnt by varying the message for creating the generations.
\item \textbf{Noise Interpolation:} Keeping the messages constant between the 2 generators we can understand the impact of the noise by interpolation between the 2 noise.
\end{itemize}

A very interesting insight emerged is that the interpolation between the messages showed major content of the image while the interpolation between the noise produced texture changes in the image. This phenomenon would be elucidated in the results and analysis section further.

\section{Results and Analysis}

\paragraph{Classification Results on SVHN}
As shown in \tabref{table:svhn} all of the models' discriminator representation improved the results over the discriminator representation of  DCGAN \cite{radford2015unsupervised} thus showing  that the proposed models provide  regularization to the training procedure of the discriminator. The non-trivial accuracy  obtained by the message representation which never got to see the real images is an interesting phenomena while the improvement of the accuracy with the Message alongside the Discriminator features shows that the  Message representation learns some complementary features which  helps in the overall  classification task.  The  conceding objective performs better than the competing objective in the absence of message passing while it  lags behind the competing objective  when  message passing is introduced as well.  The message passing in itself doesn't perform well as compared to the conditioned message passing with respect to the experiment performed on the generators getting noise from different noise distributions hence the rest of the message passing experiments were conducted with the conditioned message generator based architecture.

\begin{table}
	\centering
	\begin{tabular}{| l | l | l | l |}
		\hline 
	     Model &  Discriminator Rep  &  Message Rep   &  Msg+Disc Rep \\  \hline
		 DCGAN \cite{radford2015unsupervised} & 22.48\% & NA & NA \\ \hline         
		 Improved GANs \cite{salimans2016improved} & 8.11 $\pm$ 1.3 \% & NA & NA \\ \hline
Different Noise MP & 20.1\% & 53.48\% & 18.7\% \\ \hline
         \textbf{Different Noise CMP} & \textbf{17.1\%} & 54.21\% & \textbf{15.2\%} \\ \hline
         Conceding CMP & 18.37\% & 64.46\% & 17.4\% \\ \hline
         Competing CMP & 17.76\% & \textbf{52.05\%} & 16.8\% \\ \hline
         Competing Objective & 18.02\% & NA & NA \\ \hline
         Conceding Objective & 17.56\% & NA & NA \\ \hline
	\end{tabular}
    \vspace{-6pt}
	\caption{Error in Classification on SVHN. MP stands for message passing while CMP stands for conditioned message passing.\vspace{-14pt}}
	\label{table:svhn}
\end{table}

\paragraph{Clustering}

\begin{figure}[H]
    \centering
    \includegraphics[scale=0.45]{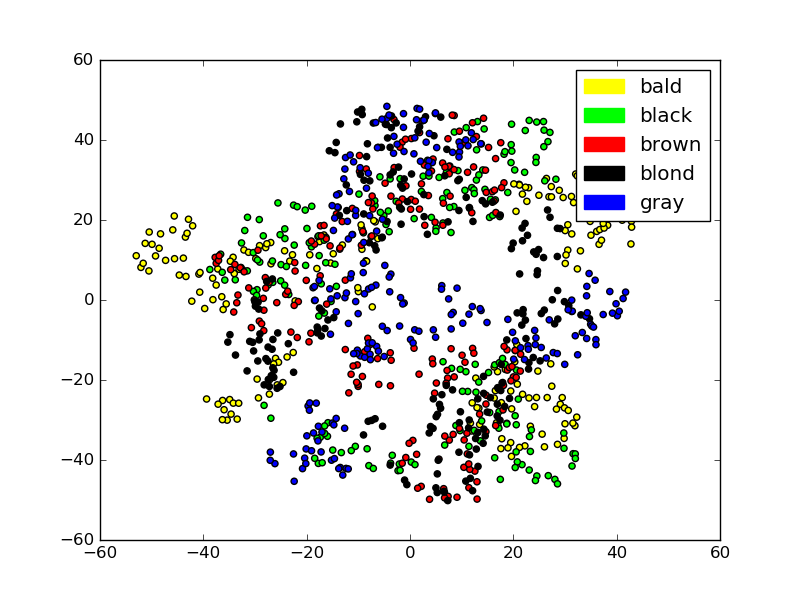}
	\caption{Clustering messages from different noise distribution with conditioned message.}
   \label{fig:cl_an}
\end{figure}
As described in the experimental section the clustering was performed with the messages and visualized using T-SNE on the 2-D space. As evident from the clustering results there emerge some clusters from the messages based on the disjoint division of the hair style. As evident from \figref{fig:cl_an} the messages for the bald hair style totally separates from the rest while black and brown being a bit subjective are similar in the message space but some clusters for black hair emerge which are totally pure. Gray hair also separates quite clearly from the rest.

\paragraph{Competing Objective} Interpolation results for competing objective.
\begin{figure}[H]
    \centering
    \includegraphics[scale=0.7]{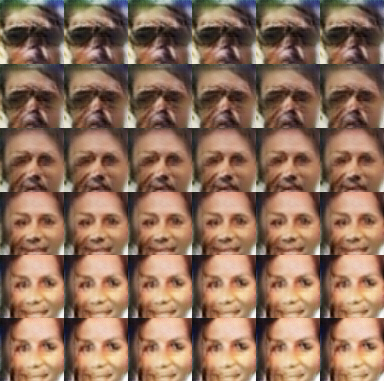}
	\caption{Noise interpolation for $G_1$ of Competing Objective. The generations obtained for noise interpolation over generator $G_1$ move from an old man wearing spectacles to a smiling lady without spectacles. The generations seem realistic. The generator is able to capture the facial details and even the direction of lightning.}
   \label{fig:inter_compG1}
\end{figure}

\begin{figure}[H]
    \centering
    \includegraphics[scale=0.7]{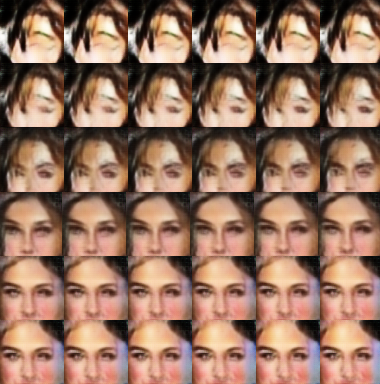}
	\caption{Noise interpolation for $G_2$ of Competing Objective. It shows that the generations by $G_2$ are more like animated characters. The generator is able to capture the dominating features but not the texture of the images. As the noise interpolation is done, the generations go from a cartoon character to its human version.}
   \label{fig:inter_compG2}
\end{figure}

\paragraph{Conceding Objective} Interpolation results for conceding objective.
\begin{figure}[H]
    \centering
    \includegraphics[scale=0.8]{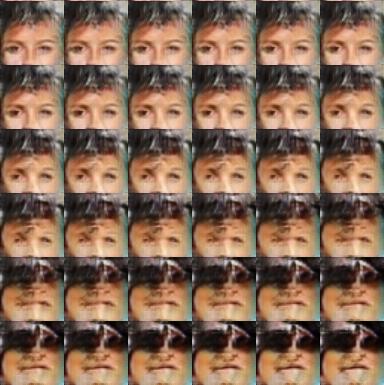}
	\caption{Noise interpolation for $G_1$ of Conceding Objective. For conceding objective, the generations go from spike hairs to normal hairs to black spectacles. Also, the face shifts us and changes to smile.}
   \label{fig:inter_concG1}
\end{figure}

\begin{figure}[H]
    \centering
    \includegraphics[scale=0.8]{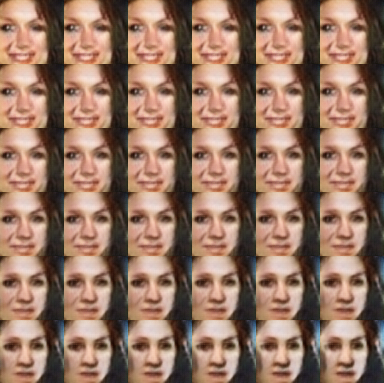}
	\caption{Noise interpolation for $G_2$ of Conceding Objective. The generated personalities are changing their mood from smile to sadness to shock. With the interpolation of noise, are also changing their orientation from straight to tilted.}
   \label{fig:inter_concG2}
\end{figure}

\paragraph{Message Passing}
We consider three bottlenecks:
\paragraph{Different Noise Distribution}

As evident from \tabref{table:svhn} that in case of different noise distribution, one with condition performs better, we consider only conditioned message passing in the next two bottlenecks:
\paragraph{Competing Objective} Interpolation results for competing objective with conditioned message.
\begin{figure}[H]
    \centering
    \includegraphics[scale=0.8]{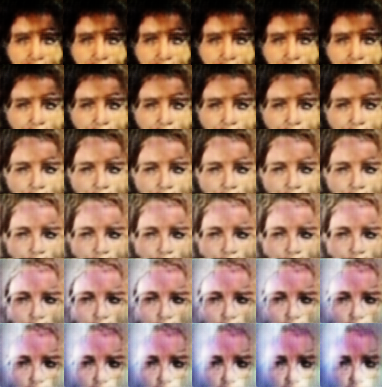}
	\caption{Noise interpolation for $G_1$ of Competing Objective with conditioned message. We see that the generator is able to learn minute details of the face and later on it is getting artistic and able to generate an angel like image with varied color schemes.}
   \label{fig:inter_comp_noisG1}
\end{figure}

\begin{figure}[H]
    \centering
    \includegraphics[scale=0.8]{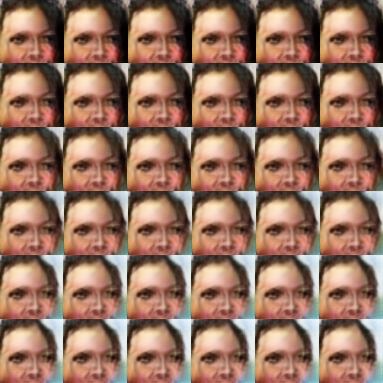}
	\caption{Noise interpolation for $G_2$ of Competing Objective with conditioned message. It's easy to see that the $G_2$ is not learning very detailed features. As the interpolation is done between the noise, the figure goes from loose hair to tied hair without making any other strong changes.}
   \label{fig:inter_comp_noisG2}
\end{figure}

\begin{figure}[H]
    \centering
    \includegraphics[scale=0.6]{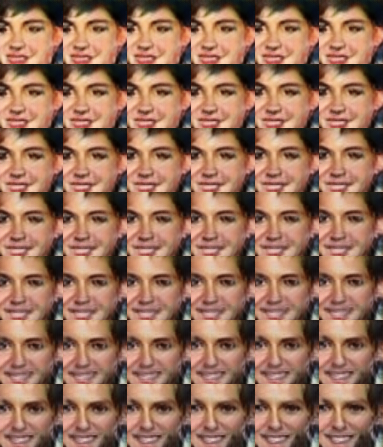}
	\caption{Message interpolation for $G_1$ of Competing Objective with conditioned message. It is made by interpolating between messages for generator $G_1$. It shows the figure changing from a smiling woman to a smiling man. The direction face is pointing to also changes.}
   \label{fig:inter_comp_msgG1}
\end{figure}

\begin{figure}[H]
    \centering
    \includegraphics[scale=0.8]{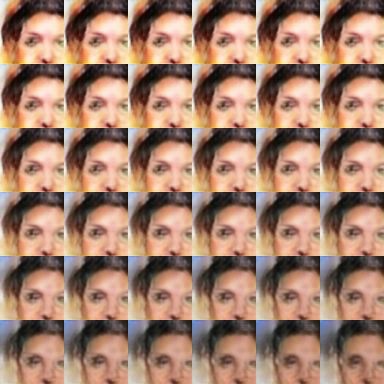}
	\caption{Message interpolation for $G_2$ of Competing Objective with conditioned message. It shows that $G_2$ hasn't learnt detailed features. The generations go from longer hair to shorter ones with changes in lightning.}
   \label{fig:inter_comp_msgG2}
\end{figure}

\paragraph{Conceding Objective} Interpolation results for conceding objective with conditioned message.
\begin{figure}[H]
    \centering
    \includegraphics[scale=0.8]{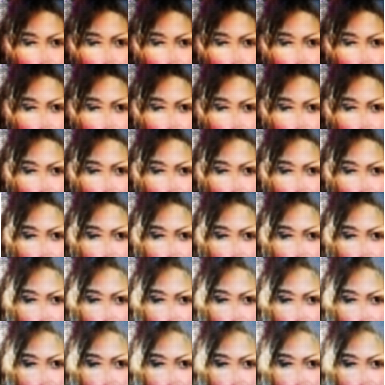}
	\caption{Noise interpolation for $G_1$ of Conceding Objective with conditioned message. With the change in noise, the lighting condition is changing. The only visible change is the appearance of face.}
   \label{fig:inter_conc_noisG1}
\end{figure}

\begin{figure}[H]
    \centering
    \includegraphics[scale=0.8]{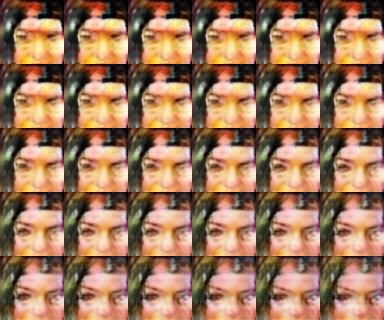}
	\caption{Noise interpolation for $G_2$ of Conceding Objective with conditioned message. We see that $G_2$ is modeling a cartoon figure with a band on forehead. With the interpolation in noise, the band is changing into hair.}
   \label{fig:inter_conc_noisG2}
\end{figure}

\begin{figure}[H]
    \centering
    \includegraphics[scale=0.8]{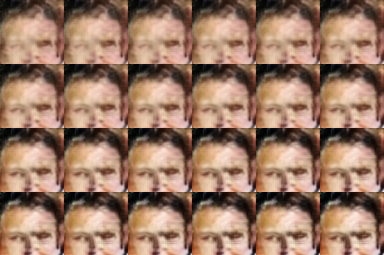}
	\caption{Message interpolation for $G_1$ of Conceding Objective with conditioned message. It seems to be the way an artist adds attributes to the face beginning with the left eye. Also the facial details get more prominent with images.}
   \label{fig:inter_conc_msgG1}
\end{figure}

\begin{figure}[H]
    \centering
    \includegraphics[scale=0.8]{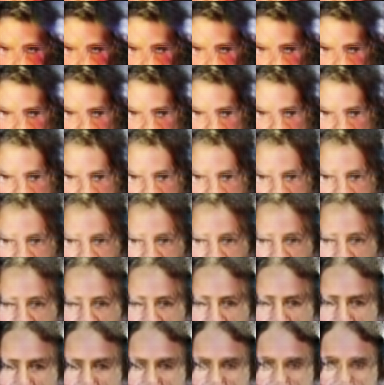}
	\caption{Message interpolation for $G_2$ of Conceding Objective with conditioned message. It shows the generations going from young person with dense hairs to old person with sparse hairs. The direction where the person is looking also changes.}
   \label{fig:inter_conc_msgG2}
\end{figure}

\section{Conclusion}

We presented several novel architectures and objectives aimed at training multi-agent GANs along with bottlenecks such as the generators receiving noise from different noise distributions, competing generators which compete with each other, conceding  generators which tries to encourage the other generator to  perform better than itself. As is evident from the experiments the models  learn meaningful representations. The  introduction of the architecture regularizes the training of the discriminator as is evident from the improved results of the discriminator.  The  representations obtained from the message generator are quite valuable in itself as is evident from the high accuracy obtained from its representation that was not even shown the real images.

{
\small
\small
\small
\small
\bibliography{biblio}

\begin{thebibliography}{}

\bibitem[Abadi and Andersen, 2016]{abadi2016learning}
Abadi, M. and Andersen, D.~G. (2016).
\newblock Learning to protect communications with adversarial neural
  cryptography.
\newblock {\em arXiv preprint arXiv:1610.06918}.

\bibitem[Arjovsky and Bottou, 2016]{arjovsky2016towards}
Arjovsky, M. and Bottou, L. (2016).
\newblock Towards principled methods for training generative adversarial
  networks.
\newblock Technical report.
\newblock Submitted to ICLR 2017.

\bibitem[Cai and Daskalakis, 2011]{cai2011minmax}
Cai, Y. and Daskalakis, C. (2011).
\newblock On minmax theorems for multiplayer games.
\newblock In {\em Proceedings of the twenty-second annual ACM-SIAM symposium on
  Discrete Algorithms}, pages 217--234. SIAM.

\bibitem[Chen et~al., 2016]{chen2016infogan}
Chen, X., Duan, Y., Houthooft, R., Schulman, J., Sutskever, I., and Abbeel, P.
  (2016).
\newblock Infogan: Interpretable representation learning by information
  maximizing generative adversarial nets.
\newblock {\em arXiv preprint arXiv:1606.03657}.

\bibitem[Dumoulin et~al., 2016]{dumoulin2016adversarially}
Dumoulin, V., Belghazi, I., Poole, B., Lamb, A., Arjovsky, M., Mastropietro,
  O., and Courville, A. (2016).
\newblock Adversarially learned inference.
\newblock {\em arXiv preprint arXiv:1606.00704}.

\bibitem[Durugkar et~al., 2016]{durugkar2016generative}
Durugkar, I., Gemp, I., and Mahadevan, S. (2016).
\newblock Generative multi-adversarial networks.
\newblock {\em arXiv preprint arXiv:1611.01673}.

\bibitem[Foerster et~al., 2016]{foerster2016learning}
Foerster, J.~N., Assael, Y.~M., de~Freitas, N., and Whiteson, S. (2016).
\newblock Learning to communicate with deep multi-agent reinforcement learning.
\newblock {\em arXiv preprint arXiv:1605.06676}.

\bibitem[Ghosh et~al., 2017]{ghosh2017contextual}
Ghosh, A., Kulharia, V., Mukerjee, A., Namboodiri, V., and Bansal, M. (2017).
\newblock Contextual rnn-gans for abstract reasoning diagram generation.
\newblock In {\em Proceedings of AAAI}.

\bibitem[Goodfellow et~al., 2014]{goodfellow2014generative}
Goodfellow, I., Pouget-Abadie, J., Mirza, M., Xu, B., Warde-Farley, D., Ozair,
  S., Courville, A., and Bengio, Y. (2014).
\newblock Generative adversarial nets.
\newblock In {\em Advances in Neural Information Processing Systems}, pages
  2672--2680.

\bibitem[Goodfellow et~al., 2013]{goodfellow2013multi}
Goodfellow, I.~J., Bulatov, Y., Ibarz, J., Arnoud, S., and Shet, V. (2013).
\newblock Multi-digit number recognition from street view imagery using deep
  convolutional neural networks.
\newblock {\em arXiv preprint arXiv:1312.6082}.

\bibitem[Kingma and Welling, 2013]{kingma2013auto}
Kingma, D.~P. and Welling, M. (2013).
\newblock Auto-encoding variational bayes.
\newblock {\em arXiv preprint arXiv:1312.6114}.

\bibitem[Lazaridou et~al., 2016]{lazaridou2016multi}
Lazaridou, A., Peysakhovich, A., and Baroni, M. (2016).
\newblock Multi-agent cooperation and the emergence of (natural) language.
\newblock Technical report.
\newblock Submitted to ICLR 2017.

\bibitem[Ledig et~al., 2016]{ledig2016photo}
Ledig, C., Theis, L., Husz{\'a}r, F., Caballero, J., Aitken, A., Tejani, A.,
  Totz, J., Wang, Z., and Shi, W. (2016).
\newblock Photo-realistic single image super-resolution using a generative
  adversarial network.
\newblock {\em arXiv preprint arXiv:1609.04802}.

\bibitem[Lee et~al., 2016]{lee2016stochastic}
Lee, S., Purushwalkam, S., Cogswell, M., Ranjan, V., Crandall, D., and Batra,
  D. (2016).
\newblock Stochastic multiple choice learning for training diverse deep
  ensembles.
\newblock {\em arXiv preprint arXiv:1606.07839}.

\bibitem[Liu and Tuzel, 2016]{liu2016coupled}
Liu, M.-Y. and Tuzel, O. (2016).
\newblock Coupled generative adversarial networks.
\newblock {\em arXiv preprint arXiv:1606.07536}.

\bibitem[Liu et~al., 2015]{liu2015faceattributes}
Liu, Z., Luo, P., Wang, X., and Tang, X. (2015).
\newblock Deep learning face attributes in the wild.
\newblock In {\em Proceedings of International Conference on Computer Vision
  (ICCV)}.

\bibitem[Makhzani et~al., 2015]{makhzani2015adversarial}
Makhzani, A., Shlens, J., Jaitly, N., and Goodfellow, I. (2015).
\newblock Adversarial autoencoders.
\newblock {\em arXiv preprint arXiv:1511.05644}.

\bibitem[Mirza and Osindero, 2014]{mirza2014conditional}
Mirza, M. and Osindero, S. (2014).
\newblock Conditional generative adversarial nets.
\newblock {\em arXiv preprint arXiv:1411.1784}.

\bibitem[Nowozin et~al., 2016]{nowozin2016f}
Nowozin, S., Cseke, B., and Tomioka, R. (2016).
\newblock f-gan: Training generative neural samplers using variational
  divergence minimization.
\newblock {\em arXiv preprint arXiv:1606.00709}.

\bibitem[Pathak et~al., 2016]{pathak2016context}
Pathak, D., Krahenbuhl, P., Donahue, J., Darrell, T., and Efros, A.~A. (2016).
\newblock Context encoders: Feature learning by inpainting.
\newblock {\em arXiv preprint arXiv:1604.07379}.

\bibitem[Radford et~al., 2015]{radford2015unsupervised}
Radford, A., Metz, L., and Chintala, S. (2015).
\newblock Unsupervised representation learning with deep convolutional
  generative adversarial networks.
\newblock {\em arXiv preprint arXiv:1511.06434}.

\bibitem[Reed et~al., 2016]{reed2016generative}
Reed, S., Akata, Z., Yan, X., Logeswaran, L., Schiele, B., and Lee, H. (2016).
\newblock Generative adversarial text to image synthesis.
\newblock {\em arXiv preprint arXiv:1605.05396}.

\bibitem[Salimans et~al., 2016]{salimans2016improved}
Salimans, T., Goodfellow, I., Zaremba, W., Cheung, V., Radford, A., and Chen,
  X. (2016).
\newblock Improved techniques for training gans.
\newblock {\em arXiv preprint arXiv:1606.03498}.

\bibitem[Sukhbaatar et~al., 2016]{sukhbaatar2016learning}
Sukhbaatar, S., Szlam, A., and Fergus, R. (2016).
\newblock Learning multiagent communication with backpropagation.
\newblock {\em arXiv preprint arXiv:1605.07736}.

\bibitem[Vondrick et~al., 2016]{vondrick2016generating}
Vondrick, C., Pirsiavash, H., and Torralba, A. (2016).
\newblock Generating videos with scene dynamics.
\newblock {\em arXiv preprint arXiv:1609.02612}.

\bibitem[Weiss and Freeman, 2001]{weiss2001optimality}
Weiss, Y. and Freeman, W.~T. (2001).
\newblock On the optimality of solutions of the max-product belief-propagation
  algorithm in arbitrary graphs.
\newblock {\em IEEE Transactions on Information Theory}, 47(2):736--744.

\bibitem[Wu et~al., 2016]{wu2016learning}
Wu, J., Zhang, C., Xue, T., Freeman, W.~T., and Tenenbaum, J.~B. (2016).
\newblock Learning a probabilistic latent space of object shapes via 3d
  generative-adversarial modeling.
\newblock {\em arXiv preprint arXiv:1610.07584}.

\bibitem[Zhu et~al., 2016]{zhu2016generative}
Zhu, J.-Y., Kr{\"a}henb{\"u}hl, P., Shechtman, E., and Efros, A.~A. (2016).
\newblock Generative visual manipulation on the natural image manifold.
\newblock In {\em European Conference on Computer Vision}, pages 597--613.
  Springer.

\end{thebibliography}
}

\bibliographystyle{apalike}

\end{document}